\documentclass[runningheads,a4paper]{llncs}

\usepackage{amssymb}
\usepackage{graphicx}
\usepackage{multirow}
\usepackage{color}
\usepackage[colorlinks,linkcolor=blue]{hyperref}
\usepackage{epstopdf}

\usepackage{url}
\urldef{\mailsa}\path|{alfred.hofmann, ursula.barth, ingrid.haas, frank.holzwarth,|
\urldef{\mailsb}\path|anna.kramer, leonie.kunz, christine.reiss, nicole.sator,|
\urldef{\mailsc}\path|erika.siebert-cole, peter.strasser, lncs}@springer.com|    
\newcommand{\keywords}[1]{\par\addvspace\baselineskip
\noindent\keywordname\enspace\ignorespaces#1}

\begin{document}
\mainmatter  
\title{Multi-Estimator Full Left Ventricle Quantification through Ensemble Learning}
\titlerunning{  }
\author{Jiasha Liu\inst{1\dagger},Xiang Li\inst{2\dagger},Hui Ren\inst{2,4\dagger},Quanzheng Li\inst{1,2,3}}
\institute{$^{1}$Center for Data Science, Peking University.\\
$^{2}$MGH/BWH Center for Clinical Data Science.\\
$^{3}$Laboratory for Biomedical Image Analysis, Beijing Institute of Big Data Research.\\
$^{4}$Heart Center, Peking University People’s Hospital.\\
$\dagger$ Joint First Authors}
\authorrunning{  }
\toctitle{Lecture Notes in Computer Science}
\tocauthor{Authors' Instructions}
\maketitle
\begin{abstract}
Cardiovascular disease accounts for 1 in every 4 deaths in United States. Accurate estimation of structural and functional cardiac parameters is crucial for both diagnosis and disease management. In this work, we develop an ensemble learning framework for more accurate and robust left ventricle (LV) quantification. The framework combines two 1st-level modules: direct estimation module and a segmentation module. The direct estimation module utilizes Convolutional Neural Network (CNN) to achieve end-to-end quantification. The CNN is trained by taking 2D cardiac images as input and cardiac parameters as output. The segmentation module utilizes a U-Net architecture for obtaining pixel-wise prediction of the epicardium and endocardium of LV from the background. The binary U-Net output is then analyzed by a separate CNN for estimating the cardiac parameters. We then employ linear regression between the 1st-level predictor and ground truth to learn a 2nd-level predictor that ensembles the results from 1st-level modules for the final estimation. Preliminary results by testing the proposed framework on the LVQuan18 dataset show superior performance of the ensemble learning model over the two base modules. 
\end{abstract}
\section{Introduction}
Cardiac MRI is a widely-adopted, accurate non-ionizing method for the assessment of cardiac disease, especially in evaluating cardiac function and morphology. Quantitative measures from short axes (SAX) cine are recommended in reporting guideline to describe left ventricular (LV) size and function, such as volume, wall thickness and regional wall motion of left ventricle \cite{RN1122}. Manually quantitative measurements are time-consuming, less reproducible and increasing the burden of radiologists. Automatic quantification of the cardiac parameters has become an increasingly important problem, especially in recent years due to the success in applying computer vision techniques on various image processing tasks. In this work, we present an ensemble learning framework which combines the results from two well-established quantification methods: direct estimation-based method and segmentation-based method, and ensembles them through a linear regression model, to further improve the prediction power. Preliminary results on LVQuan18 dataset using 3-folds cross validation scheme shows that the proposed ensemble learning framework can achieve accurate and robust left ventricle quantification for both structural and functional cardiac parameters.
\section*{Related Works}
Previous literature on cardiac parameters estimation mainly belong to two categories: the segmentation-based methods \cite{RN1076,RN1123,RN1124} which perform LV segmentation first, then measure the parameters based on segmentation results. This category of methods is advantages as the results (usually as segmentation masks) are easier to interpret which can provide more insight into the analytic procedure. The latter category of direct estimation-based methods \cite{RN1125,RN1126,RN1127,RN1128} perform estimation of the cardiac parameters directly from the image and/or image features, without relying on explicit segmentation procedure. It is advantages as the training error can be directly back-propagated to the feature selection and regression process, thus normally resulting in higher accuracy. Also, direct estimation methods do not require manual annotation of cardiac regions, thus are more feasible to be used in real practice. 
\section{Ensemble Learning Framework}
\subsection{Overview and Data Description}
In this work, we utilize two 1st-level (“base”) modules for LV quantification: a direct estimation module which is based on an end-to-end CNN structure to map the input 2D cardiac image to its quantitative measurements; and a semantic segmentation module which is based on a U-Net structure to map the same input image to a binary mask indicating its epicardium and endocardium, which are then fed into a separate CNN to obtain the measurements. In order to effectively and robustly combine them, we then train a 2nd-level ensemble classifier based on linear regression to map the results from these two modules to the ground truth measurements. The proposed framework is evaluated on cardiac MR dataset provided by LVQuan18. The dataset consists processed SAX MR sequences of 145 subject, where for each subject there are 20 frames to capture the whole cardiac cycle. Ground truth measurements as well as annotations for epicardium and endocardium are provided for every single frame.

The goal of full left ventricle quantification is estimating three types of measurement indices, including; 1) A1, A2: cavity area and myocardium area; 2) D1-D3: dimensions of cavity of three directions (AS-IL, IS-AL, and I-AL); and 3) RWT1~RWT6: regional wall thickness, staring from the anterior-septal segment in counter clockwise direction. In addition, the cardiac phase as a binary value (1 for systolic and 0 for diastolic) will also be inferred from the quantification. Evaluation (training and testing) of the proposed system is performed using 3-folds cross-validation scheme.
\subsection{Direct estimation module}
\begin{figure}
\centering
\includegraphics[width =\textwidth]{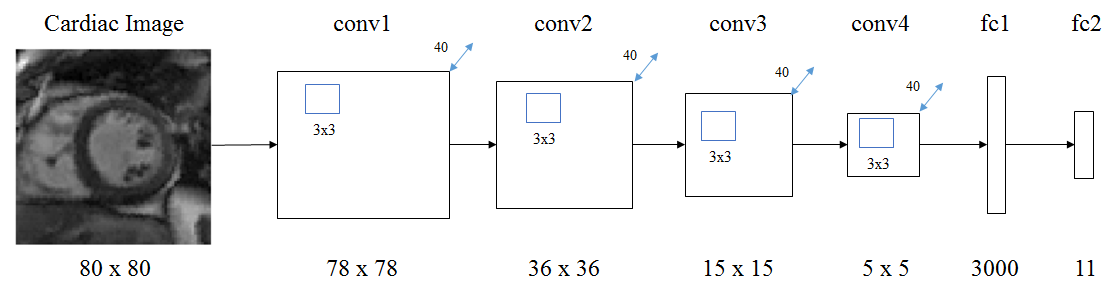}
\caption{Diagram of the direct estimation network. Each convolution layer is followed by a rectified linear unit layer (ReLU), a batch normalization layer and a max-pooling layer.}
\label{fig:Fig1}
\end{figure}

Structure of CNN in the direct estimation module consists of four convolution layers (conv1~4) and two fully-connected (fc1~2) layers, which is visualized in ~\ref{fig:Fig1}. The direct estimation module essentially performs non-linear regression between the input image (high-dimensional, pixel-wise representation of a 2-D object) and its characterizations (low-dimensional, semantic information). The effective mapping between high and low-dimensional data is achieved through convolutional filters, which perform dimension reduction and feature extraction on images based on the premise of shift-invariance, local connectivity and compositionality. 

\begin{figure}
\centering
\includegraphics[width =\textwidth]{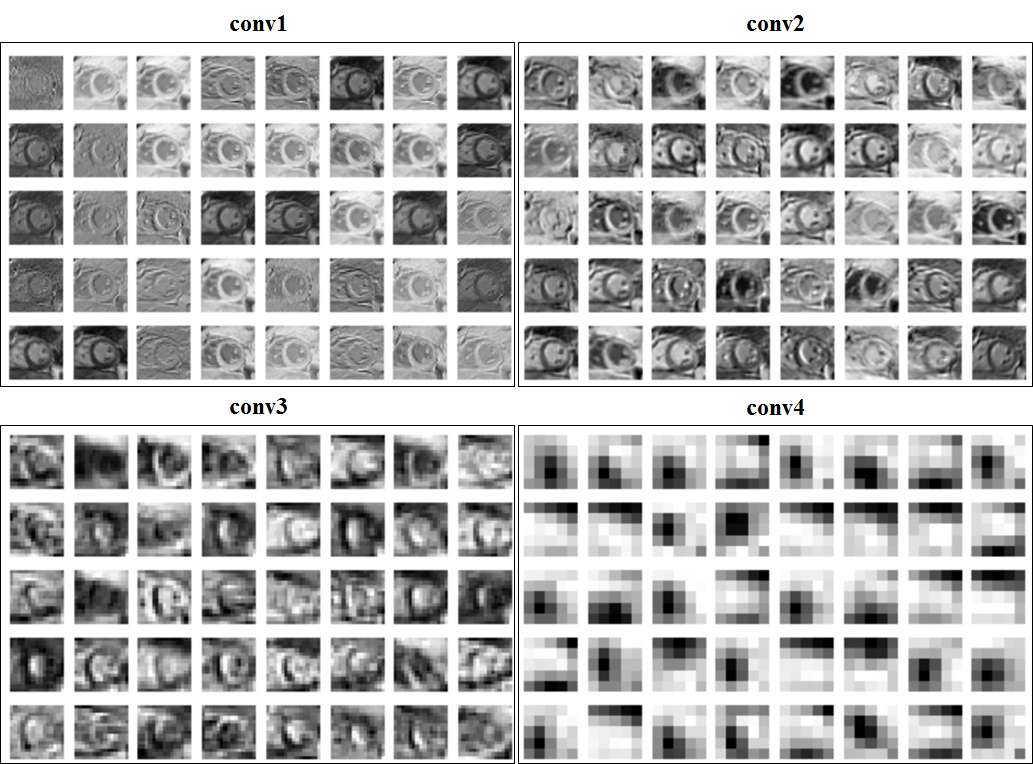}
\caption{Visualization of conv1- conv4 layers output (i.e. feature maps) of the direct estimation CNN on a sample image.}
\label{fig:Fig2}
\end{figure}

In addition, we also analyzed the outputs of each convolutional kernel in conv1-conv4 from a given image and obtained their visualizations (i.e. “feature maps”) in Fig.~\ref{fig:Fig2}. It can be observed that convolutional kernels are capable of effectively representing structure information and local variability of the input images.
\clearpage

\subsection{Segmentation module}
\begin{figure}
\centering
\includegraphics[width =\textwidth]{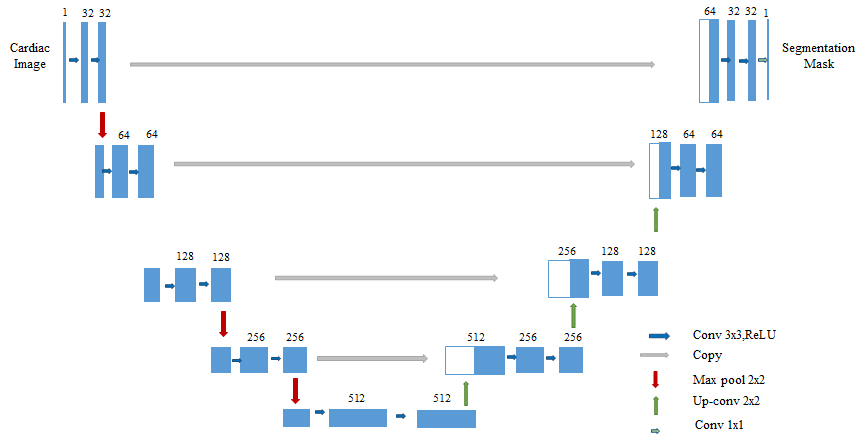}
\caption{Diagram of the U-Net architecture used in this work. Each blue box represents a multi-channel feature map. Number on top of the blue box shows number of channels. Arrows in different color correspond to different operations, as shown in the figure legends.}
\label{fig:Fig3}
\end{figure}

In segmentation module, we apply U-Net for cardiac image segmentation task, as shown in Fig.~\ref{fig:Fig3}. U-Net has been successfully applied in many visual tasks, especially in biomedical image processing. It is based on encoder-decoder architecture, where the spatial dimension is reduced by pooling layers in encoder, and the object details and spatial dimension are gradually recover in decoder \cite{RN1129}. In contrast to direct estimation module, here the input cardiac image is explicitly mapped to the manual annotation masks through U-Net. After the segmentation results (i.e. binary mask of the cardiac image) are obtained, we use a CNN with three convolution layers and two fully-connected layers to map the binary mask to the quantification measurements. 

\begin{figure}
\centering
\includegraphics[width =\textwidth]{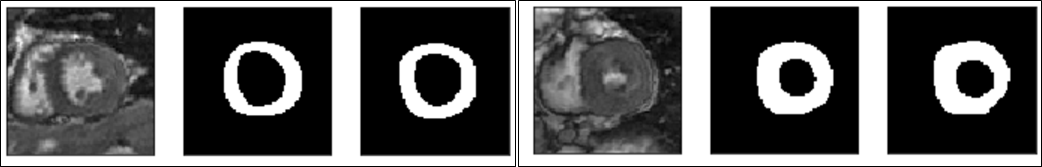}
\caption{Sample results by U-Net on two randomly-selected cardiac images. In each of the image, Left column: input cardiac images; Middle column: ground truth annotations; Right column: segmentation results.}
\label{fig:Fig4}
\end{figure}

Sample segmentation results are shown in ~\ref{fig:Fig4}, illustrating that U-Net has very robust performance on images with noises and varying cardiac structures: DICE coefficient of the segmentation reaches higher than 0.9 in the test data. In both direct estimation module and segmentation module, the networks are implemented by Keras, using Adam optimizer. Batch size and epochs are determined by cross-validation results combined with grid search.
\clearpage
\subsection{2nd-level ensemble learning from base modules}
Prediction results from the two 1st-level base modules, represented as two 11×1 vectors for each subject, are then used for training a 2nd-level ensemble learning framework. The ensemble predictor is based on linear regression between the 1st-level predictor results and ground truth measurements in the training dataset. Learned linear ensemble model thus consists of weights between the two base predictors for each quantification, which will be used to predict the final estimation in the testing dataset. Final output of the ensemble framework is the quantification measurements as one 11×1 vector for each subject. Cardiac phase is inferred separately from other measurements, which is based on a thresholding for the predicted area parameters and regularization rules (at most two change points for the systolic/ diastolic transition through the whole cycle).
\section{Experiments and Results}
In this work, we use 3-folds cross-validation scheme for performance evaluation and comparison, where the 145-subjects LVQuan18 dataset is divided into three groups with (49, 48, 48) subjects. Cross-validation performances of the 1st-level direct estimation module and the segmentation module, as well as the 2nd-level ensemble learning framework which serves as the final output of the proposed system, are summarized in the tables below. In addition, average accuracy of the threshold-based phase estimator based on the 2nd-level prediction results is 88

In Table 1, value in each cell is the MAE (mean average error) of the prediction, plus/minus standard deviation. Best prediction accuracy (i.e. lowest MAE) is highlighted in bold. It should be noted that direct estimation module and segmentation module have different levels of prediction power for different quantification. Specifically, segmentation module performs better for area and dimension estimation, which are more directly related to the LV segmentation; direct estimation module performs better for RWT (i.e. LV regional wall thicknesses) estimation which is a more complicated and abstract measurement. While the ensemble learning framework outperforms the two base modules in all fields, indicating that it successfully combines the results from two methods with different underlying mechanisms and utilizes results from weaker predictors to further improve the quantification accuracy. 
\begin{table}[]
\caption {Performance comparison between the 1st-level modules and 2nd-level ensembles in the cross-validation experiment. } 
\centering
\renewcommand{\arraystretch}{1.2}
\begin{tabular}{|c|c|c|c|}
\hline
&   Direct Estimation   & Segmentation  & Ensemble  \\ \hline    
\multicolumn{4}{|c|}{\textit{Area} (mm\textsuperscript{2})} \\ \hline
\textit{cavity} & 112.61$\pm$89.29  & 95.22$\pm$80.46   & \textbf{88.75$\pm$72.72}  \\ \hline
\textit{myocardium} & 172.97$\pm$145.39 & 163.11$\pm$119.31 & \textbf{159.50$\pm$125.14} \\ \hline
average & 142.79$\pm$117.34 & 129.17$\pm$99.88  & \textbf{124.13$\pm$98.93}  \\ \hline
\multicolumn{4}{|c|}{\textit{Dimension} (mm)}   \\ \hline
\textit{dims1}  & 2.81$\pm$2.22 & 2.37$\pm$1.91 & \textbf{2.22$\pm$1.80}    \\ \hline
\textit{dims2}  & 2.62$\pm$2.12 & 2.52$\pm$2.03 & \textbf{2.28$\pm$1.87}    \\\hline
\textit{dims3}  & 2.82$\pm$2.29 & 2.38$\pm$1.93 & \textbf{2.32$\pm$1.81}    \\\hline
average & 2.75$\pm$2.21 & 2.56$\pm$0.96 & \textbf{2.27$\pm$1.83} \\ \hline
\multicolumn{4}{|c|}{\textit{RWT} (mm)}   \\ \hline
\textit{IS} & 1.68$\pm$1.41 & 1.82$\pm$1.33 & \textbf{1.64$\pm$1.32} \\ \hline
\textit{I}  & 1.65$\pm$1.31 & 1.62$\pm$1.22 & \textbf{1.57$\pm$1.22} \\\hline
\textit{IL} & 1.90$\pm$1.61 & 1.90$\pm$1.50 & \textbf{1.83$\pm$1.49} \\\hline
\textit{AL} & 1.76$\pm$1.50 & 1.79$\pm$1.38 & \textbf{1.69$\pm$1.39} \\ \hline
\textit{A}  & 1.42$\pm$1.29 & 1.54$\pm$1.20 & \textbf{1.41$\pm$1.22} \\\hline
\textit{AS} & 1.65$\pm$1.32 & 1.60$\pm$1.26 & \textbf{1.56$\pm$1.25} \\\hline
average & 1.68$\pm$1.41 & 1.71$\pm$1.31  & \textbf{1.62$\pm$1.32} \\ \hline
\end{tabular}
\end{table}

\begin{figure}
\centering
\includegraphics[width =\textwidth]{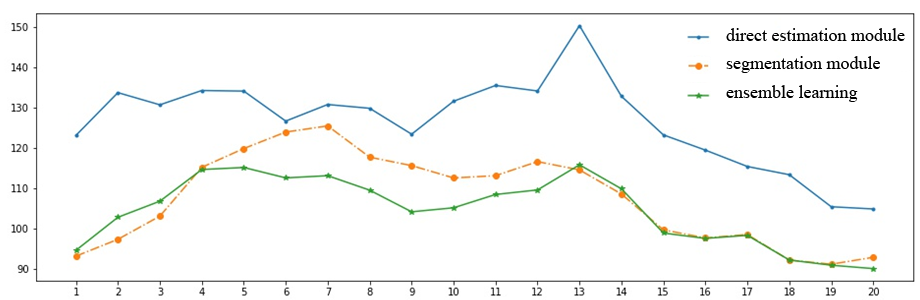}
\caption{Average frame-wise estimation errors of the area quantification estimated by the two 1st-level module (direct estimation in blue and segmentation in orange), as well as the ensemble learning framework (green). X-axis shows the indices of temporal frames, from 1st frame to the 20th frame. Y-axis shows the MSE of predictions, measured in mm\textsuperscript{2}.}
\label{fig:Fig5}
\end{figure}

In addition to the cross-validation experiments, we also obtain the average MAEs of the prediction in different time frames. Visualizations for the temporal changes of the prediction performance, using area parameters estimation as example, are shown in ~\ref{fig:Fig5}. From the temporal analysis it can be observed that our ensemble learning framework can obtain smoother estimation for the quantification, reducing impacts from frames with large prediction errors, which indicates that by ensemble predictions from multiple sources, we can improve the model robustness on more difficult/complex instances. 
\clearpage
\section{Conclusions}
In this work, we propose a multi-estimator, ensemble learning-based framework for full LV quantification. Preliminary results on the LVQuan18 dataset shows that the proposed framework can outperform the commonly applied quantification methods by integrating predictions from multiple models in a learning-based approach. Currently only two 1st-level base modules are included in the framework, in our next step of investigation we are aiming to employ more quantification methods, especially the unsupervised learning-based image processing algorithms, into our framework to further improve its robustness in different image settings. 
\bibliographystyle{unsrt}
\bibliography{reference.bib}

\begin{thebibliography}{1}

\bibitem{RN1122}
W.~Gregory Hundley, David Bluemke, Jan~G. Bogaert, Matthias~G. Friedrich,
  Charles~B. Higgins, Mark~A. Lawson, Michael~V. McConnell, Subha~V. Raman,
  Albert~C. van Rossum, Scott Flamm, Christopher~M. Kramer, Eike Nagel, and
  Stefan Neubauer.
\newblock Society for cardiovascular magnetic resonance guidelines for
  reporting cardiovascular magnetic resonance examinations.
\newblock {\em Journal of Cardiovascular Magnetic Resonance}, 11(1):5, 2009.

\bibitem{RN1076}
Peng Peng, Karim Lekadir, Ali Gooya, Ling Shao, Steffen~E. Petersen, and
  Alejandro~F. Frangi.
\newblock A review of heart chamber segmentation for structural and functional
  analysis using cardiac magnetic resonance imaging.
\newblock {\em Magnetic Resonance Materials in Physics, Biology and Medicine},
  29(2):155--195, 2016.

\bibitem{RN1123}
Ismail Ben~Ayed, Hua-mei Chen, Kumaradevan Punithakumar, Ian Ross, and Shuo Li.
\newblock Max-flow segmentation of the left ventricle by recovering
  subject-specific distributions via a bound of the bhattacharyya measure.
\newblock {\em Medical Image Analysis}, 16(1):87--100, 2012.

\bibitem{RN1124}
Caroline Petitjean and Jean-Nicolas Dacher.
\newblock A review of segmentation methods in short axis cardiac mr images.
\newblock {\em Medical Image Analysis}, 15(2):169--184, 2011.

\bibitem{RN1125}
M.~Afshin, I.~B. Ayed, K.~Punithakumar, M.~Law, A.~Islam, A.~Goela, T.~Peters,
  and S.~Li.
\newblock Regional assessment of cardiac left ventricular myocardial function
  via mri statistical features.
\newblock {\em IEEE Transactions on Medical Imaging}, 33(2):481--494, 2014.

\bibitem{RN1126}
Z.~Wang, M.~B. Salah, B.~Gu, A.~Islam, A.~Goela, and S.~Li.
\newblock Direct estimation of cardiac biventricular volumes with an adapted
  bayesian formulation.
\newblock {\em IEEE Transactions on Biomedical Engineering}, 61(4):1251--1260,
  2014.

\bibitem{RN1127}
Wufeng Xue, Andrea Lum, Ashley Mercado, Mark Landis, James Warrington, and Shuo
  Li.
\newblock Full quantification of left ventricle via deep multitask learning
  network respecting intra- and inter-task relatedness.
\newblock In Maxime Descoteaux, Lena Maier-Hein, Alfred Franz, Pierre Jannin,
  D.~Louis Collins, and Simon Duchesne, editors, {\em Medical Image Computing
  and Computer-Assisted Intervention − MICCAI 2017}, pages 276--284. Springer
  International Publishing.

\bibitem{RN1128}
W.~Xue, A.~Islam, M.~Bhaduri, and S.~Li.
\newblock Direct multitype cardiac indices estimation via joint representation
  and regression learning.
\newblock {\em IEEE Transactions on Medical Imaging}, 36(10):2057--2067, 2017.

\bibitem{RN1129}
Olaf Ronneberger, Philipp Fischer, and Thomas Brox.
\newblock U-net: Convolutional networks for biomedical image segmentation.
\newblock In Nassir Navab, Joachim Hornegger, William~M. Wells, and
  Alejandro~F. Frangi, editors, {\em Medical Image Computing and
  Computer-Assisted Intervention – MICCAI 2015}, pages 234--241. Springer
  International Publishing.

\end{thebibliography}
\clearpage
\end{document}